\documentclass[a4paper]{llncs}

\usepackage[utf8]{inputenc}

\usepackage{pdflscape}
\usepackage{booktabs}
\usepackage{amsmath}
\usepackage{mathtools}
\usepackage{multirow}
\usepackage[final]{pdfpages}
\usepackage{algorithm}
\usepackage[noend]{algpseudocode}
\floatname{algorithm}{Procedure}

\usepackage{caption}
\usepackage{subcaption}
\usepackage{adjustbox}
\usepackage{amsfonts}
\usepackage{wrapfig}
\usepackage{array}
\usepackage{tabularx}

\usepackage[english]{babel}

\usepackage[%
rm={oldstyle=false,proportional=true},%
sf={oldstyle=false,proportional=true},%
tt={oldstyle=false,proportional=true,variable=true},%
qt=false%
]{cfr-lm}
\usepackage{graphicx}

\usepackage{cite}

\usepackage[T1]{fontenc}

\usepackage[math]{blindtext}

\usepackage{csquotes}

\usepackage{microtype}

\usepackage{url}
\makeatletter
\g@addto@macro{\UrlBreaks}{\UrlOrds}
\makeatother

\usepackage{xcolor}

\usepackage[
bookmarks=false,
breaklinks=true,
colorlinks=true,
linkcolor=black,
citecolor=black,
urlcolor=black,
pdfpagelayout=SinglePage,
pdfstartview=Fit
]{hyperref}
\usepackage[all]{hypcap}

\usepackage{pdfcomment}

\newcommand{\rakel}{RA$k$EL } 
\newcommand{\rakeld}{RA$k$EL$d$ }

\usepackage[capitalise,nameinlink]{cleveref}
\crefname{section}{Sect.}{Sect.}
\Crefname{section}{Section}{Sections}

\usepackage{xspace}

\DeclareFontFamily{U}{MnSymbolC}{}
\DeclareSymbolFont{MnSyC}{U}{MnSymbolC}{m}{n}
\DeclareFontShape{U}{MnSymbolC}{m}{n}{
    <-6>  MnSymbolC5
   <6-7>  MnSymbolC6
   <7-8>  MnSymbolC7
   <8-9>  MnSymbolC8
   <9-10> MnSymbolC9
  <10-12> MnSymbolC10
  <12->   MnSymbolC12%
}{}
\DeclareMathSymbol{\powerset}{\mathord}{MnSyC}{180}

\begin{document}

\input glyphtounicode.tex
\pdfgentounicode=1

\title{Is a Data-Driven Approach still Better than Random Choice with Naive Bayes classifiers?}

\author{Piotr Szymański \and Tomasz Kajdanowicz}
\institute{Department of Computational Intelligence, Wrocław University of Technology, Wybrzeże Stanisława Wyspiańskiego 27, 50-370 Wrocław, Poland}

%

\maketitle

\begin{abstract}
We study the performance of data-driven, a priori and random approaches to label space partitioning for multi-label classification with a Gaussian Naive Bayes classifier. Experiments were performed on 12 benchmark data sets and evaluated on 5 established measures of classification quality: micro and macro averaged F1 score, subset accuracy and Hamming loss. Data-driven methods are significantly better than an average run of the random baseline. In case of F1 scores and Subset Accuracy - data driven approaches were more likely to perform better than random approaches than otherwise in the worst case. There always exists a method that performs better than a priori methods in the worst case. The advantage of data-driven methods against a priori methods with a weak classifier is lesser than when tree classifiers are used.
\end{abstract}


\section{Introduction}\label{sec:intro}

In our recent work \cite{DBLP:journals/entropy/SzymanskiKK16} we proposed a data-driven community detection approach to partition the label space for the multi-label classification as an alternative to random partitioning into equal subsets as performed by the random $k$-label sets method proposed by Tsoumakas et. al. \cite{Tsoumakas:2007:Rakel}. The data-driven approach works as follows: we construct a label co-occurrence graph (both weighted and unweighted versions) based on training data and perform community detection to partition the label set. Then, each partition constitutes a label space for separate multi-label classification sub-problems. As a result, we obtain an ensemble of multi-label classifiers that jointly covers the whole label space. We consider a variety of approaches:  modularity-maximizing techniques approximated by fast greedy and leading eigenvector methods, infomap, walktrap and label propagation algorithms. For comparison purposes we evaluate the binary relevance (BR) and label powerset (LP) - which we call a priori methods, as they a priori assume a total partitioning of the label space into singletons (BR) and lack of any partitioning (LP).

The variant of \rakel evaluated in this paper is an approach in which the label space is either partitioned into equal-sized subsets of labels. This approach is called \rakeld - \rakel distinct as the label sets are non-overlapping. \rakeld takes one parameter - the number of label sets to partition into $k$. We assumed that all partitions are equally probable and that the remainder of the label set smaller than $k$ becomes the last element of the otherwise equally sized partition family.

In \cite{DBLP:journals/entropy/SzymanskiKK16} we compared community detection methods to label space divisions against \rakeld and a priori methods on 12 benchmark datasets (\textit{bibtex} \cite{katakis_multilabel_2008}, \textit{delicious} \cite{tsoumakas_effective_2008}, \textit{tmc2007} \cite{tsoumakas_effective_2008}, \textit{enron} (\cite{klimt2004enron}), \textit{medical} \cite{read_classifier_2011}, \textit{scene} \cite{boutell_learning_2004}, \textit{birds} \cite{birds},  \textit{Corel5k} \cite{duygulu_object_2002}, \textit{Mediamill} \cite{snoek_challenge_2006}, \textit{emotions} \cite{trohidis2008multi}, \textit{yeast} \cite{elisseeff_kernel_2001}, \textit{genbase} \cite{Diplaris2005}) over five evaluation measures with Classifier and Regression Trees (CART) as base classifiers. We discovered that data-driven approaches are more efficient and more likely to outperform \rakeld than binary relevance or label powerset is, in every evaluated measure. For all measures, apart from Hamming loss, data-driven approaches are significantly better than RAkELd ($\alpha=0.05$ ), and at least one data-driven approach is more likely to outperform RAkELd than a priori methods in the case of RAkELd’s best performance. This has been the largest RAkELd evaluation published to date with 250 samplings per value for 10 values of RAkELd parameter k on 12 datasets published to date.

In this paper we extend our result and evaluate whether the same results hold if instead of using tree-based methods, we employ a weak and Gaussian Naive Bayesian classifier from the scikit-learn python package \cite{scikit-learn}. The experimental setup remains identical to the one presented in tree-based scheme, except for the change of base classifier.  Bayesian classifiers remain of interest in many applications due to their low computational requirements.

We thus repeat the research questions we have asked in the case of tree-based classifiers, this time for Naive Bayes based classifiers:
\begin{itemize}
\item[] \textbf{RH1}: Data-driven approach is significantly better than random ($\alpha$ = 0.05)  
\item[] \textbf{RH2}: Data-driven approach is more likely to outperform RAkELd than a priori methods
\item[] \textbf{RH3}: Data-driven approach is more likely to outperform RAkELd than a priori methods in the worst case
\item[] \textbf{RH4}: Data-driven approach is more likely to perform better than RAkELd in the worst case, than otherwise
\end{itemize}

\section{Results}

\begin{table}[b]
    \centering
    \small

\begin{tabular}{llllll}
\toprule
& FG & FGW & LE & LEW & WTW \\
\midrule
Macro-averaged F1 & 0.068  & 0.37 & 0.054 & 0.37 & 0.37 \\
Micro-averaged F1 &{\bf 0.011} & 0.071 & {\bf 0.003}  &{\bf 0.011} &{\bf 0.043} \\
Jaccard Score &  {\bf 0.026} & 0.07 & {\bf 0.008} &  {\bf 0.026} &  0.070 \\   

\bottomrule
\end{tabular}

\label{tab:pvalues}
\caption{P-values of data-driven methods performing better than an average run of RA$k$EL$d$ for each measure tested using non-parametric Friedman test with Rom's post-hoc test. Only methods with p-values greater than $\alpha=0.05$ are presented. All approaches not listed explicitly were significantly better than \rakeld in all measures.}
\end{table}

\paragraph{Micro-averaged F1 score.} While a priori methods such as Binary Relevance and Label Powerset exhibit a higher median likelihood of outperforming \rakeld - we note that the highest mean likelihood is obtained by label propagation data-driven label space division on an unweighted label co-occurrence graph. Unweighted label propagation is also most likely to outperform \rakeld in the worst case. Thus we reject \textbf{RH2} and accept \textbf{RH3} and \textbf{RH4}. The best performing and recommended community detection method for micro-averaged F1 score - unweighted label propagation - is better than average performance of \rakeld with statistical significance, we thus accept \textbf{RH1}.

\begin{table}[b]
    \centering
    \begin{adjustbox}{width=1\textwidth}
    \small

\begin{tabular}{lrrrr}
\toprule
{} &   Minimum &    Median &      Mean &       Std \\
\midrule
BR                           &  0.369565 &  \textbf{1.000000} &  0.796143 &  0.283117 \\
LP                           &  0.369565 &  0.999076 &  0.789076 &  0.294146 \\
fastgreedy                   &  0.263556 &  0.781778 &  0.737634 &  0.232243 \\
fastgreedy-weighted          &  0.322667 &  0.601848 &  0.633698 &  0.160196 \\
infomap                      &  0.448000 &  0.869778 &  0.817113 &  0.194957 \\
infomap-weighted             &  0.091556 &  0.797333 &  0.705199 &  0.299230 \\
label\_propagation            &  \textbf{0.529778} &  0.908000 &  \textbf{0.843744} &  0.172125 \\
label\_propagation-weighted   &  0.317778 &  0.662356 &  0.703653 &  0.243097 \\
leading\_eigenvector          &  0.302667 &  0.829778 &  0.748593 &  0.250929 \\
leading\_eigenvector-weighted &  0.341778 &  0.632063 &  0.684237 &  0.185325 \\
walktrap                     &  0.321333 &  0.717391 &  0.719968 &  0.246686 \\
walktrap-weighted            &  0.239556 &  0.600889 &  0.632683 &  0.221396 \\
\bottomrule
\end{tabular}

\end{adjustbox}
\label{tab:f1-micro.all}
\caption{Likelihood of performing better than RA$k$EL$d$ in Micro-averaged F1 score of every method for each data set}
\end{table} 


\begin{table}[h]
    \centering
    \begin{adjustbox}{width=1\textwidth}

\begin{tabular}{lrrrrrrrrrrrr}
\toprule
{} &        BR &        LP &  FG &  FGW  &  IN &  INW &  LPG &  LPGW &  LE &  LEW &  WT &  WTW \\
\midrule
Corel5k     &  0.39 &  0.37 &    0.85 &             0.79 &  0.87 &          0.09 &           0.99 &                    0.32 &             0.9 &                      0.91 &  0.43 &           0.68 \\
bibtex      &  0 &  0 &    0.26 &             0.32 &  0.45 &          0.3 &           0.53 &                    0.34 &             0.30 &                      0.34 &  0.32 &           0.24 \\
birds       &  0 &  0.999 &    0.62 &             0.6 &  0.66 &          0.66 &           0.66 &                    0.66 &             0.79 &                      0.62 &  0.95 &           0.34 \\
delicious   &  0 &  0 &    0.78  &             0.59 &  0.87 &          0.59 &           0.87 &                    0.62 &             0.83 &                      0.72 &  0.54 &           0.58 \\
emotions    &  0.43 &  0.37 &    0 &             0.52 &  0 &          0 &           0 &                    0.57 &             0 &                      0.52 &  0 &           0.57 \\
enron       &  0.98 &  0.98 &    0.94 &             0.88 &  0.93 &          0.93 &           0.93 &                    0.93 &             0 &                      0.99 &  0.79 &           0.997 \\
mediamill   &  0 &  0 &    0.55 &             0.65 &  0.91 &          0.8 &           0.91 &                    0.91 &             0.45 &                      0.69 &  0.68 &           0.6 \\
medical     &  0 &  0 &    0.51 &             0.58 &  0.51 &          0.60 &           0.60 &                    0.60 &             0.41 &                      0.59 &  0.51 &           0.60 \\
scene       &  0.37 &  0.37 &    0.72 &             0.63 &  0.80 &          0.80 &           0.80 &                    0.80 &             0.72 &                      0.63 &  0.72 &           0.63 \\
tmc2007-500 &  0 &  0 &    0.89 &             0.55 &  0 &          0 &           0 &                    0 &             0.85 &                      0.63 &  0 &           0.89 \\
yeast       &  0.58 &  0.59 &    0.99 &             0.85 &  0.99 &          0.99 &           0.99 &                    0.99 &             0.99 &                      0.88 &  0.99 &           0.83 \\
\bottomrule
\end{tabular}

\end{adjustbox}
\label{tab:f1-micro.all1}
\caption{Likelihood of performing better than RA$k$EL$d$ in Micro-averaged F1 score of every method for each data set. BR - Binary Relevance, LP - Label Powerset, FG - fastgreedy, FGW - fastgreedy weighted, IN - infomap, INW - infomap weighted, LPG - label propagation, LPGW - label propagation weighted, LE - leading eigenvector, LEW - leading eigenvector weighted, WT - walktrap, WTW - walktrap weighted.}
\end{table} 

\paragraph{Macro-averaged F1 score.} In case of macro averaged F1 score Label Powerset is the most likely to outperform \rakeld both in median and mean cases, while underperforms in the worst case. Label propagation data-driven label space division on an unweighted label co-occurrence graph is the most likely data-driven approach to outperform \rakeld - although other approaches also yield good results. Unweighted label propagation is also most likely to outperform \rakeld in the worst case. It is also better than an average run of \rakeld with statistical significance. Thus we accept \textbf{RH1}, reject \textbf{RH2} and accept \textbf{RH3} and \textbf{RH4}.

\begin{table}[b]
    \centering
    \begin{adjustbox}{width=1\textwidth}
    \small

\begin{tabular}{lrrrr}
\toprule
{} &   Minimum &    Median &      Mean &       Std \\
\midrule
BR                           &  0.456522 &  \textbf{1.000000} &  \textbf{0.868708} &  0.222246 \\
LP                           &  0.434783 &  \textbf{1.000000} &  0.850310 &  0.227355 \\
fastgreedy                   &  0.376444 &  0.836000 &  0.799503 &  0.210402 \\
fastgreedy-weighted          &  0.378222 &  0.753333 &  0.679727 &  0.175535 \\
infomap                      &  \textbf{0.519630} &  0.806861 &  0.810572 &  0.164820 \\
infomap-weighted             &  0.188444 &  0.739130 &  0.728628 &  0.247947 \\
label\_propagation            &  \textbf{0.519630} &  0.878667 &  0.841961 &  0.163304 \\
label\_propagation-weighted   &  0.500000 &  0.739130 &  0.751203 &  0.186984 \\
leading\_eigenvector          &  0.367111 &  0.806861 &  0.746465 &  0.232450 \\
leading\_eigenvector-weighted &  0.358667 &  0.832457 &  0.722748 &  0.215705 \\
walktrap                     &  0.253778 &  0.877333 &  0.789586 &  0.225409 \\
walktrap-weighted            &  0.302222 &  0.800444 &  0.745813 &  0.235022 \\
\bottomrule
\end{tabular}

\end{adjustbox}
\label{tab:f1-macro.all}
\caption{Likelihood of performing better than RA$k$EL$d$ in Macro-averaged F1 score of every method for each data set}
\end{table} 


\begin{table}[ht]
    \centering
    \begin{adjustbox}{width=1\textwidth}
    \small

\begin{tabular}{lrrrrrrrrrrrr}
\toprule
{} &        BR &        LP &  FG &  FGW  &  IN &  INW &  LPG &  LPGW &  LE &  LEW &  WT &  WTW \\
\midrule
Corel5k     &  0.94 &  0.78 &    0.37 &             0.37 &  0.89 &          0.18 &           0.997 &                    0.76 &             0.36 &                      0.36 &  0.25  &           0.3  \\
bibtex      &  1.0 &  1.0 &    0.53 &             0.57 &  0.67 &          0.52 &           0.88 &                    0.55 &             0.52 &                      0.61  &  0.6  &           0.47 \\
birds       &  1.0 &  1.0 &    0.98 &             0.84 &  0.52 &          0.52 &           0.52 &                    0.52 &             0.99 &                      0.96 &  0.97 &           0.97 \\
delicious   &  1.0  &  1.0  &    1.0  &             0.79  &  1.0  &          1.0  &           1.0  &                    1.0  &             1.0  &                      0.85  &  0.88  &           0.97  \\
emotions    &  0.46 &  0.46 &    0.93  &             0.52  &  0.93  &          0.93  &           0.93  &                    0.5 &             0.93 &                      0.52  &  0.93  &           0.5 \\
enron       &  1.0  &  0.998 &    0.986 &             0.89  &  0.66  &          0.66 &           0.66  &                    0.66 &             0.88  &                      0.89  &  0.99  &           0.91 \\
mediamill   &  1.0  &  1.0  &    0.84  &             0.75  &  0.99  &          0.91  &           0.99  &                    0.99  &             0.76  &                      0.84  &  0.89 &           0.8  \\
medical     &  1.0  &  1.0 &    0.7 &             0.45 &  0.7  &          0.74 &           0.74 &                    0.74 &             0.39  &                      0.45 &  0.70  &           0.74 \\
scene       &  0.46 &  0.43 &    0.65  &             0.65  &  0.74  &          0.74 &           0.74 &                    0.74 &             0.65 &                      0.65  &  0.65  &           0.65  \\
tmc2007-500 &  1.0 &  1.0 &    0.99  &             0.8 &  1.0 &          1.0 &           1.0 &                    1.0 &             0.92 &                      0.83  &  1.0 &           0.98 \\
yeast       &  0.7  &  0.68 &    0.8  &             0.83  &  0.81 &          0.81  &           0.81  &                    0.81  &             0.81  &                      1.0 &  0.81  &           0.91  \\
\bottomrule
\end{tabular}

\end{adjustbox}
\label{tab:f1-macro.all1}
\caption{Likelihood of performing better than RA$k$EL$d$ in Macro-averaged F1 score of every method for each data set. BR - Binary Relevance, LP - Label Powerset, FG - fastgreedy, FGW - fastgreedy weighted, IN - infomap, INW - infomap weighted, LPG - label propagation, LPGW - label propagation weighted, LE - leading eigenvector, LEW - leading eigenvector weighted, WT - walktrap, WTW - walktrap weighted.}
\end{table}

\paragraph{Subset Accuracy.} In case of Subset Accuracy label propagation performed on an unweighted graph approach to dividing the labels space is the most resilient approach both in the worst case and in the average (mean/median) likelihood. The weighted version performers equally well in the worst case, so does unweighted infomap. As the worst case performance of three data-driven methods is greater than $0.5$ we accept \textbf{RH4} for Subset Accuracy. While Label Powerset performs better than label propagation in case of the median/mean likelihood of being better than \rakeld - it performs worse by 12 pp. in the worst case. Thus while rejecting \textbf{RH2} and accepting \textbf{RH3} we still recommend using data-driven label propagation approach instead of Label Powerset. Label propagation performs better than \rakeld with statistical significance - we accept \textbf{RH1}.
\begin{table}[b]
    \centering
    \begin{adjustbox}{width=1\textwidth}
    \small

\begin{tabular}{lrrrr}
\toprule
{} &   Minimum &    Median &      Mean &       Std \\
\midrule
BR                           &  0.217391 &  0.886667 &  0.777640 &  0.285316 \\
LP                           &  0.413043 &  \textbf{1.000000} &  \textbf{0.924946} &  0.174772 \\
fastgreedy                   &  0.028637 &  0.585333 &  0.621030 &  0.304067 \\
fastgreedy-weighted          &  0.007852 &  0.586728 &  0.512003 &  0.225171 \\
infomap                      &  0.429000 &  0.978500 &  0.887924 &  0.203588 \\
infomap-weighted             &  \textbf{0.533487} &  0.934783 &  0.831424 &  0.195409 \\
label\_propagation            &  \textbf{0.533487} &  0.998222 &  0.912394 &  0.165066 \\
label\_propagation-weighted   &  \textbf{0.533487} &  0.934783 &  0.834437 &  0.180916 \\
leading\_eigenvector          &  0.000000 &  0.644000 &  0.604389 &  0.355451 \\
leading\_eigenvector-weighted &  0.000000 &  0.568988 &  0.499787 &  0.304284 \\
walktrap                     &  0.133487 &  0.600000 &  0.625201 &  0.295569 \\
walktrap-weighted            &  0.000000 &  0.608696 &  0.499824 &  0.331589 \\
\bottomrule
\end{tabular}

\end{adjustbox}
\label{tab:accuracy.all}
\caption{Likelihood of performing better than RA$k$EL$d$ in Subset Accuracy of every method for each data set}
\end{table}

\begin{table}[h]
    \centering
    \begin{adjustbox}{width=1\textwidth}
    \small

\begin{tabular}{lrrrrrrrrrrrr}
\toprule
{} &        BR &        LP &  FG &  FGW  &  IN &  INW &  LPG &  LPGW &  LE &  LEW &  WT &  WTW \\
\midrule
Corel5k     &  0.34 &  0.87 &    0.59 & 0.68 &  0.99 & 0.59 & 0.998 &                    0.83  &             0.0  &                      0.0 &  0.34  &           0.0 \\
bibtex      &  0.89 &  1.0 &    0.37 &             0.69 &  0.96  &          0.61 &           0.998 &                    0.7 &             0.64 &                      0.73 &  0.37 &           0.31 \\
birds       &  0.996 &  0.997 &    0.029  &             0.007 &  0.53 &          0.53 &           0.53 &                    0.53 &             0.09 &                      0.0 &  0.13 &           0.03 \\
delicious   &  1.0 &  1.0 &    0.997  &             0.63  &  1.0 &          0.999  &           1.0 &                    1.0 &             1.0 &                      0.79  &  0.79  &           0.92 \\
emotions    &  0.21  &  0.41  &    1.0 &             0.48  &  1.0 &          1.0 &           1.0 &                    0.61 &             1.0 &                      0.48 &  1.0 &           0.61  \\
enron       &  0.86  &  1.0 &    0.58  &             0.57  &  0.98 &          0.98 &           0.98 &                    0.98 &             0.79 &                      0.67  &  0.6 &           0.65 \\
mediamill   &  1.0 &  1.0 &    0.45  &             0.29  &  0.96  &          0.87 &           0.96 &                    0.96 &             0.41 &                      0.38 &  0.57 &           0.28  \\
medical     &  1.0 &  1.0 &    0.43 &             0.64  &  0.43 &          0.64 &           0.64 &                    0.64 &             0.33 &                      0.65  &  0.43 &           0.64 \\
scene       &  0.63 &  0.93 &    0.63 &             0.3 &  0.93 &          0.93 &           0.93 &                    0.93 &             0.63 &                      0.3 &  0.63 &           0.3 \\
tmc2007-500 &  1.0 &  1.0 &    0.75  &             0.59 &  1.0 &          1.0 &           1.0 &                    1.0 &             0.75  &                      0.57 &  1.0 &           0.87  \\
yeast       &  0.62 &  0.96 &    0.999 &             0.76  &  0.999 &          0.999 &           0.999 &                    0.999 &             0.999 &                      0.92  &  0.999 &           0.88 \\
\bottomrule
\end{tabular}

\end{adjustbox}
\label{tab:accuracy.all1}
\caption{Likelihood of performing better than RA$k$EL$d$ in Subset Accuracy of every method for each data set. BR - Binary Relevance, LP - Label Powerset, FG - fastgreedy, FGW - fastgreedy weighted, IN - infomap, INW - infomap weighted, LPG - label propagation, LPGW - label propagation weighted, LE - leading eigenvector, LEW - leading eigenvector weighted, WT - walktrap, WTW - walktrap weighted.}
\end{table} 

\paragraph{Jaccard score.} Among data-driven methods the label propagation performed on an unweighted graph approach to dividing the labels space is the most resilient approach both in the worst case and in the average (mean/median) likelihood. It is followed by infomap. While a priori methods are perform better in case of the median likelihood by 3 pp., they perform worse than data-driven methods in the mean and worst case. We thus confirm \textbf{RH2} and \textbf{RH3}. The worst case likelihood of data-driven methods outperforming \rakeld is not grater than $0.5$ we thus reject \textbf{RH4}. Unweighted infomap performs better than the average run of \rakeld with statistical significance - we thus accept \textbf{RH1}.


\begin{table}[b]
    \centering
    \begin{adjustbox}{width=1\textwidth}
    \small

\begin{tabular}{lrrrr}
\toprule
{} &   Minimum &    Median &      Mean &       Std \\
\midrule
BR                           &  0.326087 &  \textbf{1.000000} &  0.784597 &  0.303331 \\
LP                           &  0.369565 &  \textbf{1.000000} &  0.847350 &  0.240611 \\
fastgreedy                   &  0.183372 &  0.756000 &  0.674557 &  0.274675 \\
fastgreedy-weighted          &  0.177367 &  0.586957 &  0.591697 &  0.194144 \\
infomap                      &  \textbf{0.411085} &  0.925333 &  0.831944 &  0.218665 \\
infomap-weighted             &  0.053778 &  0.804889 &  0.686328 &  0.327207 \\
label\_propagation            &  \textbf{0.411085} &  0.974500 & \textbf{ 0.86552} &  0.203504 \\
label\_propagation-weighted   &  0.239111 &  0.630435 &  0.689132 &  0.281967 \\
leading\_eigenvector          &  0.308000 &  0.777333 &  0.693396 &  0.272005 \\
leading\_eigenvector-weighted &  0.116859 &  0.653745 &  0.624674 &  0.222935 \\
walktrap                     &  0.359556 &  0.696444 &  0.668188 &  0.252658 \\
walktrap-weighted            &  0.080370 &  0.586957 &  0.580375 &  0.244502 \\
\bottomrule
\end{tabular}

\end{adjustbox}
\label{tab:jaccard.all}
\caption{Likelihood of performing better than RA$k$EL$d$ in Jaccard Similarity of every method for each data set}
\end{table} 

\begin{table}[h]
 \centering
 \begin{adjustbox}{width=1\textwidth}
 \small

\begin{tabular}{lrrrrrrrrrrrr}
\toprule
{} &  BR &  LP &  FG &  FGW  &  IN &  INW &  LPG &  LPGW &  LE &  LEW &  WT &  WTW \\
\midrule
Corel5k     &  0.35 &  0.47 &    0.76 & 0.8 &  0.9 &          0.05 &           0.996 & 0.24 & 0.78 &                      0.83 &  0.43 &           0.57 \\
bibtex      &  1.0 &  1.0 &    0.31  & 0.42  &  0.86 & 0.40  & 0.99 & 0.45 & 0.42 & 0.45 &  0.36 & 0.28  \\
birds       &  1.0 &  0.999  &    0.18  & 0.18 &  0.41 & 0.41 & 0.41 & 0.41  & 0.32  & 0.12  &  0.45 & 0.08 \\
delicious   &  1.0 &  1.0 &    0.7 & 0.53 &  0.77 & 0.44 &           0.77 & 0.47  & 0.74  & 0.7  &  0.49  &           0.49  \\
emotions    &  0.33 &  0.37 &    0.98 & 0.52 &  0.98 &          0.98 &           0.98 &                    0.63 &             0.98 &                      0.52 &  0.98 &           0.63 \\
enron       &  0.993  &  1.0 &    0.84 &             0.82 &  0.97  &          0.97  &           0.97  &                    0.97  &             0.999 &                      0.87  &  0.74  &           0.88  \\
mediamill   &  1.0 &  1.0 &    0.54  &             0.65  &  0.93 &          0.80  &           0.93 &                    0.93  &             0.44 &                      0.68 &  0.7 &           0.6 \\
medical     &  1.0 &  1.0 &    0.41 &             0.55 &  0.41 &          0.55 &           0.55 &                    0.55 &             0.31 &                      0.56 &  0.41 &           0.55 \\
scene       &  0.39 &  0.85 &    0.80 &             0.59 &  0.93 &          0.93 &           0.93 &                    0.93 &             0.8 &                      0.59 &  0.8 &           0.59 \\
tmc2007-500 &  1.0 &  1.0 &    0.89 &             0.6 &  1.0 &          1.0 &           1.0 &                    1.0 &             0.85 &                      0.65 &  1.0 &           0.9 \\
yeast       &  0.57  &  0.63 &    0.994 &             0.85 &  0.994 &          0.994 &           0.994 &                    0.994 &             0.994 &                      0.91 &  0.994 & 0.82 \\
\bottomrule
\end{tabular}

\end{adjustbox}
\label{tab:jaccard.all1}
\caption{Likelihood of performing better than RA$k$EL$d$ in Jaccard Similarity of every method for each data set}
\end{table}

\paragraph{Hamming Loss} The data-driven methods that are most likely to outperform \rakeld are infomap and label propagation performed on a weighted label co-occurence graph. We recommend using weighted infomap which is also most resilient in the worst case, although much less resilient than the desired $0.5$ likelihood of outperforming \rakeld in the worst case. As a result the case of Hamming Loss we confirm \textbf{RH2} and \textbf{RH3} but reject \textbf{RH4}. Weighted infomap perform significantly better than an average run of \rakeld - we accept \textbf{RH1}.

\begin{table}[b]
    \centering
    \begin{adjustbox}{width=1\textwidth}
    \small

\begin{tabular}{lrrrr}
\toprule
{} &   Minimum &    Median &      Mean &       Std \\
\midrule
BR                           &  0.110667 &  0.558538 &  0.579872 &  0.376954 \\
LP                           &  0.080889 &  0.652174 &  0.592830 &  0.379345 \\
fastgreedy                   & \textbf{ 0.208889} &  0.418222 &  0.513625 &  0.276367 \\
fastgreedy-weighted          &  0.111111 &  0.260870 &  0.302981 &  0.223065 \\
infomap                      &  0.112889 &  0.735111 &  0.684758 &  0.292563 \\
infomap-weighted             &  \textbf{0.204889} &  \textbf{0.847826} & \textbf{ 0.727799} &  0.291282 \\
label\_propagation            &  0.111111 &  0.735111 &  0.684971 &  0.312656 \\
label\_propagation-weighted   &  0.237778 &  0.735111 &  0.714660 &  0.237049 \\
leading\_eigenvector          &  0.121333 &  0.498029 &  0.552381 &  0.315482 \\
leading\_eigenvector-weighted &  0.111111 &  0.260870 &  0.337735 &  0.227415 \\
walktrap                     &  0.111111 &  0.418667 &  0.541611 &  0.331449 \\
walktrap-weighted            &  0.094226 &  0.328113 &  0.387505 &  0.228658 \\
\bottomrule
\end{tabular}

\end{adjustbox}
\label{tab:hl.all}
\caption{Likelihood of performing better than RA$k$EL$d$ in Hamming Loss of every method for each data set}
\end{table} 

\begin{table}[h]
    \centering
    \begin{adjustbox}{width=1\textwidth}
    \small

\begin{tabular}{lrrrrrrrrrrrr}
\toprule
{} &        BR &        LP &  FG &  FGW  &  IN &  INW &  LPG &  LPGW &  LE &  LEW &  WT &  WTW \\
\midrule
Corel5k     &  0.11 &  0.15 &    0.42 &  0.35 &  0.33 & 0.96 & 0.23 & 0.86 & 0.21 & 0.22 &  0.42 & 0.42 \\
bibtex      &  0.11 &  0.08 &    0.31 & 0.24 &  0.11 & 0.20 & 0.11 & 0.24 & 0.27 & 0.24 &  0.17 & 0.26 \\
birds       &  1.0 &  0.99 &    0.27 & 0.16  &  0.7 & 0.7 & 0.7 & 0.7 & 0.37 & 0.2 &  0.41 & 0.09 \\
delicious   &  0.11 &  0.11 &    0.34 & 0.11 &  0.36 & 0.24 & 0.36 & 0.39 & 0.41 & 0.11 &  0.11 & 0.2 \\
emotions    &  0.43 &  0.30 &    1.0 & 0.28 &  1.0 & 1.0 & 1.0 & 0.54 & 1.0 & 0.28 &  1.0 & 0.54 \\
enron       &  0.40 &  0.69 &    0.31 & 0.3 &  0.73 & 0.73 & 0.73 & 0.73 & 0.94 & 0.57 &  0.27 & 0.43 \\
mediamill   &  0.998 &  0.999 &    0.21 & 0.23 &  0.74 & 0.51 & 0.74 & 0.74 & 0.12 & 0.31 &  0.35 & 0.22 \\
medical     &  1.0 &  1.0 &    0.77  & 0.94 &  0.77 & 0.88 & 0.88 & 0.88 & 0.79 & 0.93 &  0.77  & 0.88 \\
scene       &  0.65 &  0.65 &    0.52 & 0.26 &  0.85 & 0.85 & 0.85 & 0.85 & 0.52 & 0.26 &  0.52 & 0.26 \\
tmc2007-500 &  1.0 &  1.0 &    0.57 & 0.17 &  1.0 & 1.0 & 1.0 & 1.0 & 0.5 & 0.26 &  1.0 & 0.64 \\
yeast       &  0.56 &  0.55 &    0.94 & 0.3 &  0.94 & 0.94 & 0.94 & 0.94 & 0.94 & 0.33 &  0.94 & 0.33 \\
\bottomrule
\end{tabular}

\end{adjustbox}
\label{tab:hl.all1}
\caption{Likelihood of performing better than RA$k$EL$d$ in Hamming Loss of every method for each data set. BR - Binary Relevance, LP - Label Powerset, FG - fastgreedy, FGW - fastgreedy weighted, IN - infomap, INW - infomap weighted, LPG - label propagation, LPGW - label propagation weighted, LE - leading eigenvector, LEW - leading eigenvector weighted, WT - walktrap, WTW - walktrap weighted.}
\end{table}

\section{Conclusion and Outlook}

We have examined the performance of data-driven, a priori and random approaches to label space partitioning for multi-label classification with a Gaussian Naive Bayes classifier. Experiments were performed on 12 benchmark data sets and evaluated on 5 established measures of classification quality. Table \ref{tab:finalrh} summarizes out findings. Data-driven methods are significantly better than an average \rakeld run that had not undergone parameter estimation - i.e. when results are compared against the mean result of all evaluated \rakeld paramater values. When compared against the likelihood of outperforming a \rakeld in the evaluated parameter space - in case of F1 scores and Subset Accuracy - data driven approaches were more likely to perform better than \rakeld than otherwise in the worst case. There always exists a method that performs better than a priori methods in the worst case. 

Data driven methods perform better than a priori methods in the mean likelihood but worse in median when it comes to micro-averaged F1 and Subset Accuracy. This can be attributed to differences in how likelihoods per data set distribute - data-driven methods perform better in worst case, but are also less likely to be always better than \rakeld as opposed to a priori methods. The advantage of data-driven methods against a priori methods with a weak classifier is lesser than when tree classifiers are used. The authors acknowledge support from the National Science Centre research projects decision no. 2016/21/N/ST6/02382 and 2016/21/D/ST6/02948.

\begin{table}[H]
\begin{tabularx}{\textwidth}{X m{1.5cm} m{1.5cm} m{1.5cm} m{1.5cm} m{1.5cm}}
    \toprule
    & Micro-averaged F1                          & Macro-averaged F1     & Subset Accuracy    & Jaccard Similarity & Hamming Loss          \\ 
    \midrule
    \textbf{RH1}                          & Yes                                        & Yes                   & Yes                & Yes                & Yes                    \\
    \textbf{RH2} & Undecided                                        & No                    & No                & Undecided                & Yes                    \\
    \textbf{RH3} & Yes                                        & Yes                   & Yes                & Yes                & Yes                   \\
    \textbf{RH4}               & Yes                                        & Yes                   & Yes                & No                & No                    \\
    \midrule
    Recommended data-driven approach  & Unweighted label propagation & Unweighted label propagation & Unweighted label propagation & Unweighted label propagation & Weighted infomap  \\
\bottomrule
\end{tabularx}
\caption{The summary of evaluated hypotheses and proposed recommendations of this paper\label{tab:finalrh}}
\end{table}

\bibliographystyle{splncs03}
\bibliography{paper}

\end{document}